# Morphemes Without Borders: Evaluating Root–Pattern Morphology in Arabic Tokenizers and LLMs

**Yara Alakeel**[1], **Chatrine Qwaider**[2], **Hanan Aldarmaki**[2], **Sawsan Alqahtani**[3,1]*

[1] Saudi Data & AI Authority (SDAIA),
[2] Mohamed bin Zayed University of Artificial Intelligence (MBZUAI)
[3] Princess Nourah Bint Abdulrahman University (PNU)
yalakeel@ncai.gov.sa, {chatrine.qwaider,hanan.aldarmaki}@mbzuai.ac.ae, saalqhtani@pnu.edu.sa

## Abstract

This work investigates how effectively large language models (LLMs) and their tokenization schemes represent and generate Arabic root-pattern morphology, probing whether they capture genuine morphological structure or rely on surface memorization. Arabic morphological system provides a rich testbed for analyzing how LLMs handle complex, non-concatenative forms and how tokenization choices influence this process. Our study begins with an evaluation of morphological fidelity across Arabic and multilingual tokenizers against gold-standard segmentation, followed by an analysis of LLM performance in productive root-pattern generation using a newly developed test set. Our findings across seven Arabic-centric and multilingual LLMs and their respective tokenizers reveal that tokenizer morphological alignment is not necessary nor sufficient for morphological generation, which questions the role of morphological tokenization in downstream performance.

**Keywords:** Arabic, non-concatenative, morphology, tokenization, LLMs, roots and patterns

## 1. Introduction

Large language models (LLMs) have achieved impressive performance across many natural language processing tasks, yet their success is uneven across languages (Grattafiori et al., 2024; Hui et al., 2025; Achiam et al., 2024). Prior work suggests that typological complexity is one factor contributing to these disparities, with evidence that LLMs tend to perform worse on morphologically rich languages, partly due to tokenization inefficiency and data sparsity (Hofmann et al., 2025). Fertility is often used as a measure of tokenization effectiveness, with higher scores assumed to be generally worse both for performance (Ali et al., 2024) and cost (Petrov et al., 2023) considerations. Languages with non-concatenative morphology, such as Arabic, pose particular challenges for current language models, which are based on contiguous tokenization schemes (Beesley and Karttunen, 2003; Habash, 2010).

Arabic morphology operates through a root-and-pattern system in which consonantal roots combine with templatic vowel patterns to produce derivational and inflectional forms. For example, applying the pattern مفعول /mfʕuːl/ to the root كتب /ktb/ yields the passive participle مكتوب /mktuːb/, where the prefix م /m/ and infix و /uː/ are inserted into the root structure to form the word. The root encodes core lexical meaning (e.g., /ktb/ 'write'), while the pattern contributes morpho-syntactic and lexical information through templatic material that may include prefixes, infixes, or suffixes.

While frequent word forms may be memorized, genuinely productive morphological learning requires that models identify roots and apply interleaving patterns to generalize to unseen forms (Ismayilzada et al., 2025; Hofmann et al., 2025). Prior studies have attempted to incorporate morphological information into LLMs, typically through morphology-based tokenization or architectural adjustments. Yet their effects remain limited, and their implications for non-concatenative languages like Arabic are still not well understood (Jabbar, 2024; Gazit et al., 2025; Asgari et al., 2025).

In this paper, we investigate how pre-trained LLMs and their tokenizers handle Arabic morphology, focusing on two key dimensions: tokenizer morphological alignment, and LLMs root-pattern morphological generation. Our goal is to understand how existing tokenization schemes align with Arabic morphological structures, and how this influences LLMs' ability to generalize productively. We focus not on efficiency or compression, but on the linguistic adequacy and representational behavior of tokenizers, examining how they support or hinder novel word construction in morphologically rich settings. Specifically, we ask: (1) to what extent do current tokenizers preserve Arabic morphological patterns, and (2) do morphological tokenizers correlate positively with the LLMs' morphological generation performance? Our contributions can be summarized as follows:[1]

1. We thoroughly evaluate the morphological alignment of various Arabic-centric and multilingual LLM tokenizers against ground-truth segmentation, focusing on both morpheme boundaries and morpheme integrity.
2. We construct a dataset for probing Arabic root-pattern morphological productivity using both real and nonce roots combined with various patterns or morphemes.
3. We evaluate instruction-tuned LLMs based on the various tokenizers in (1) and compare their performance in the morphological productivity task in (2).

---

*Corresponding author

[1]The dataset and code are available at https://github.com/YaraAlakeel/morphems_without_borders.

Surprisingly, we find that tokenizer morphological alignment does not predict effective morphological generation. Models whose tokenizers poorly align with true morpheme boundaries can still perform competitively, while models based on morphological tokenization can fail to generalize. In fact, the second highest scoring model, GPT4, has the highest fertility score and low morphological alignment scores on Arabic text, both of which are presumed to have negative impact on performance, which raises new questions about the role of tokenization in downstream performance.

## 2. Related Work

Current LLMs typically employ surface-based subword tokenizers such as BPE (Gage, 1994; Sennrich et al., 2016), UnigramLM (Kudo, 2018), and WordPiece (Schuster and Nakajima, 2012; Wu et al., 2016). These methods balance character- and word-level representations, reduce vocabulary size, and efficiently handle out-of-vocabulary items. BPE, in particular, remains widely adopted for its simplicity and scalability (Achiam et al., 2024; Grattafiori et al., 2024; Jiang et al., 2023; Hui et al., 2025). However, numerous studies show that such tokenizers often misalign with morphological structure in languages like Turkish and Korean, where subword segmentation frequently distorts morpheme boundaries and captures only partial linguistic regularities (Schmidt et al., 2025; Pagnoni et al., 2024; Rust et al., 2021; Mielke et al., 2021; Bostrom and Durrett, 2020).

**Morphology-Aware Tokenizers for Arabic:** To address this, several linguistically informed tokenization methods have been proposed to better capture the morphology of Arabic and other morphologically rich languages. Yet, their empirical benefits remain insufficiently explored. MorphBPE (Asgari et al., 2025) extends BPE with linguistic supervision, using Arabic Treebank and Farasa segmentations to align merges with morpheme boundaries. Many other tokenization schemes have been proposed to handle concatenative (Hofmann et al., 2022; Jabbar, 2024) and non-conatenative (Gazit et al., 2025) morphology. Despite this, these approaches remain largely unused in large-scale LLM training, with MorphBPE being the only one integrated into practice, specifically in the Fanar model's tokenizer (Abbas et al., 2025).

**Morphological Generalization:** Evaluating morphological generalization in LLMs requires both structural and behavioral assessments. Structural metrics such as the Morphological Alignment Score (MAS) (Abbas et al., 2025) quantify token–morpheme correspondence through sequence alignment. While informative, MAS conflates partial overlaps and boundary mismatches, which reduces interpretability. Similarly, MorphScore (Asgari et al., 2025) measures morpheme boundary accuracy but limits comparisons to two morphemes per word and functions primarily as a recall-based measure, sometimes rewarding spurious or linguistically implausible segmentations. Beyond alignment-based metrics, behavioral probing tasks complement these approaches by testing whether LLMs can use morphological structure productively and differentiate between related morphological patterns (Ismayilzada et al., 2025; Weissweiler et al., 2025).

**Effects of Morphological Complexity on LLM Performance.** A growing body of work investigates how morphological complexity influences LLM performance (Arnett and Bergen, 2025; Ismayilzada et al., 2025; Seo et al., 2025). Findings remain mixed; some studies report degraded performance on morphologically rich languages such as Turkish, attributing this to tokenization inefficiency and data sparsity rather than inadequate morphological modeling (Arnett and Bergen, 2025; Seo et al., 2025). Others show that even state-of-the-art models struggle with compositional generalization, particularly when exposed to novel roots or high morphological complexity (Ismayilzada et al., 2025). Complementary evidence from Arnett et al. (2025) suggests that morphological alignment explains little variance in downstream performance, indicating that morpheme-level alignment alone is not a reliable proxy for tokenization quality. Large-scale multilingual studies further corroborate these limitations: while LLMs handle surface morphology reasonably well, they fail to generalize productively across languages, especially for irregular or low-frequency forms (Hofmann et al., 2025; Dang et al., 2024; Weissweiler et al., 2023). Ismayilzada et al. (2025) underscore this gap through *productivity* and *systematicity* tasks involving nonce roots in Turkish and Finnish.

Building on this line of work, we analyze how various LLMs handle Arabic morphology, evaluating morphological alignment and integrity in tokenization as well as morphological productivity as a down-stream task. We evaluate both morphological and purely statisitcal tokenizers from Arabic-centric and multilingual LLMs.

## 3. Methodology

### 3.1. Investigated Tokenizers and LLMs

We evaluate tokenizers from both multilingual and Arabic-centric large language models. The multilingual set includes GPT-4 (Achiam et al., 2024), GPT-4o (Hurst et al., 2024), LLaMA 3 (Grattafiori et al., 2024), Qwen 3 (Hui et al., 2025) and Cohere (Ahmadian et al., 2025). The Arabic-centric models include Fanar (Abbas et al., 2025) and ALlam (Bari et al., 2025).[2] These tokenizers differ in their training data coverage, vocabulary construction strategies, and architectural design. Among them, Fanar is the only model whose tokenizer explicitly integrates morphemic information during development, as discussed in §2.

We maintain consistent configuration settings across all runs and generations. For every task, we utilized the same prompt across models as a result of an extensive prompt engineering phase in which we chose the best performing prompt. For the temperature, we tried values in the range of 0.0 to 0.6; we used 0.6 as multiple generations produced more diverse, non-deterministic outputs. Models such as GPT and ALLaM tend to follow the prompt instructions and provide only the desired word; so, their `max_token` is set to 8. For other models, whose typical outputs are much longer, `max_token` was set to 80.

### 3.2. Tokenizer-Morphology Alignment

**Datasets.** We evaluate on two Arabic corpora with gold-standard morphological segmentation: (1) the Arabic Treebank Part 3, or ATB3 for short (LDC2010T08) (Maamouri et al., 2010), covering Modern Standard Arabic, and (2) the BOLT Egyptian Arabic corpus (LDC2021T12) (Maamouri et al., 2021), covering a colloquial dialectal variety. They capture both formal and spoken registers of Arabic morphology. We remove diacritics to match undiacritized Arabic text used in real-world settings. This choice simplifies token matching but makes some templatic contrasts, those expressed only by short vowels or gemination, unobservable. We removed punctuation marks, numerical tokens, and English characters from both corpora, retaining only Arabic words for morphological evaluation. The resulting statistics are shown in Table 1.

**Morphological Analyzers.** To establish linguistically grounded baselines, we include two morphology-guided segmenters that explicitly model Arabic word structure. From *CAMeL Tools* (Obeid et al., 2020), we use the maximum-likelihood (MLE) segmenter (CAMEL), which combines lexicon- and rule-based heuristics with probabilistic scoring.[3] We additionally include the Farasa segmenter (Darwish and Mubarak, 2016), a fast, supervised system widely adopted in Arabic NLP. We use only their segmentation outputs and treat these analyzers as morphology-aware preprocessing baselines for subword tokenization.

| Dataset | #Sents | #Words | #Tokens | AvgTok/Sent |
|---|---|---|---|---|
| ATB3 | 12,626 | 337,312 | 571,449 | 45.3 |
| ATB3_c | 12,587 | 292,552 | 526,745 | 41.85 |
| BOLT | 19,994 | 149,940 | 223,326 | 11.2 |
| BOLT_c | 19,453 | 128,271 | 201,668 | 10.37 |

Table 1: Statistics of the evaluation datasets before and after cleaning (removal of numbers, punctuation, and English characters, denoted by '_c').

#### 3.2.1. Alignment Metrics

We quantify how closely tokenizer outputs align with Arabic morphological structure. Our evaluation targets two complementary aspects: (i) *concatenative alignment*, which measures correspondence between token and morpheme boundaries, and (ii) *morphological integrity*, which assesses the preservation of whole morphemes within token boundaries.

Let $W$ be the set of words. For each word $w \in W$, let $T(w)$ denote the sequence of predicted tokens, $G(w)$ the gold-standard morpheme segmentation, $B(w)$ and $\hat{B}(w)$ the sets of gold and predicted morpheme-internal boundaries, and $M(w)$ and $\hat{M}(w)$ the sets of gold and predicted morpheme spans, respectively.

**Fertility.** Following Rust et al. (2021), fertility measures the average number of tokens per word:

$$\text{Fert} = \frac{1}{|W|} \sum_{w \in W} |T(w)|$$

Lower fertility indicates more compact segmentation and higher compression rate.

**Morpheme Boundary Precision and Recall.** Boundary metrics evaluate the alignment between predicted and gold morpheme boundaries:

$$\text{BRecall} = \frac{\sum_w |B(w) \cap \hat{B}(w)|}{\sum_w |B(w)|},$$

---

[2] https://huggingface.co/Xenova/gpt-4
https://huggingface.co/Xenova/gpt-4o
https://huggingface.co/meta-llama/Meta-Llama-3-8B
https://huggingface.co/Qwen/Qwen3-8B
https://huggingface.co/CohereLabs/c4ai-command-r7b-12-2024
https://huggingface.co/QCRI/Fanar-1-9B
https://huggingface.co/humain-ai/ALLaM-7B-Instruct-preview

[3] We also experimented with the BERT version segmenter, but observed comparable performance to the MLE-based version.

$$\text{BPrecision} = \frac{\sum_w |B(w) \cap \hat{B}(w)|}{\sum_w |\hat{B}(w)|}$$

Boundary recall captures how many true morpheme breaks are recovered, while boundary precision penalizes spurious splits. Their harmonic mean defines the **Boundary F1** score, a balance between coverage and linguistic validity.

**Morpheme F1.** This stricter metric assesses complete morpheme recovery by requiring both start and end boundaries to match:

$$\text{F1} = \frac{1}{|W|} \sum_w \frac{2 \cdot |M(w) \cap \hat{M}(w)|}{|M(w)| + |\hat{M}(w)|}$$

Unlike boundary-level measures, Morpheme F1 rewards only fully correct morpheme spans.

**Morpheme Coverage Rate (MCR).** MCR quantifies the proportion of gold morphemes preserved as intact units within tokens. More concretely, we calculate MCR as follows:

$$\frac{1}{|W|} \sum_{w \in W} \frac{\left| \left\{ m \in M(w) : \exists \hat{m} \in \hat{M}(w),\, m \subseteq \hat{m} \right\} \right|}{|M(w)|}$$

A high MCR indicates that tokens maintain internal morphological integrity, while a low value reflects fragmentation across tokens.

**Interpretation.** Together, these metrics capture complementary aspects of morphological sensitivity. Boundary-based measures reflect segmentation granularity, whereas morpheme-level scores evaluate linguistic coherence. MCR provide interpretable upper bound on morphological faithfulness, allowing fine-grained comparison between analyzers and data-driven tokenizers.

## 3.3. Morphological Productivity Tasks

To evaluate whether language models capture the systematic mechanisms of Arabic word formation, we design a set of controlled morphological productivity tasks targeting three core capacities: (1) *pattern transformation*, reflecting non-concatenative root-pattern mapping; (2) *morpheme attachment*, capturing concatenative affixal composition; and (3) *generalization to unseen forms*, testing morphological productivity on nonce roots. These tasks serve as a diagnostic of linguistic generalization, evaluating whether a model can apply morphological rules productively rather than relying on memorized lexical associations (Weissweiler et al., 2025; Ismayilzada et al., 2025).

### 3.3.1. Dataset Construction

We designed a dataset to enable controlled evaluation of Arabic derivational morphology across multiple tasks. The dataset integrates both attested and synthetic (nonce) examples to assess model generalization rather than mere morphological memorization. The real root subset is derived from the Arabic Billion Words corpus[4], and the words were first processed using CAMEL to extract their roots and patterns, from which affixal information (prefixes and suffixes) was identified. Overall, this subset contains 13 different patterns and 130 unique root-pattern forms. Further, each root–pattern pair appears in three surface forms: one unaffixed and two affixed variants. The examples were all manually selected and verified to ensure correctness. The nonce subset provides a controlled test for generalization beyond seen words, as the resulting words are non-existent but are morphologically valid. The set consists of 20 synthetic roots generated randomly from arabic alphabet characters and manually checked by native speakers to ensure they do not correspond to any valid roots in Arabic. Each root was manually combined with five different patterns, resulting in 100 unique combinations. An example of a real entry from our dataset is shown below.

```
{
"root": "ثمر",
"template": "فعال",
"base_form": "ثمار",
"prefix": "ال",
"suffix": " ",
"full_form": "الثمار",
"has_affix": "true",
"root_category": "high_frequency"
}
```

Figure 1: Example dataset instance for the pattern فعال /fiʕaːl/. The instance shows the root ثمر /θmr/, for the full form الثمار /alθimaːr/. Morphological features (e.g., root, pattern) were obtained using CAMEL Tools, from which other fields (e.g., affixes and template) were identified.

### 3.3.2. Task Definition

We implement a prompt-based generation framework comprising two main conditions, each defined by the morphological information provided to the model, enabling targeted evaluation of its morphological knowledge. We test both Arabic and English prompts under 0-shot and 1-shot settings. We evaluate root-pattern and concatenative morphology

---
[4] https://huggingface.co/datasets/oserikov/arabic_billion_words

(affixation) separately and in combination More specifically, we define the following tasks:

**Root-Pattern:** The model receives a triliteral root and a derivational pattern ("`template`") and must generate the corresponding derived form ("`base_form`"). We test set with both real roots and nonce roots. The English prompt is shown below:

> In Arabic, words are formed by applying a morphological pattern to a triliteral root. Each root consists of three consonants and follows the abstract root pattern فَعَل (fa'ala).
>
> Given the root {root} and the target morphological pattern {template}, generate the corresponding Arabic word by correctly applying the root to the specified pattern.
>
> Respond with only the fully-formed Arabic word—no transliteration, spaces, punctuation, or explanation.

**Affix-Build:** the model receives the un-affixed `base_word` and an unordered set of affixes and must produce the correctly ordered derived `full_form`. This tasks test whether the model captures concatenative affix attachment and ordering.

> Arabic Unaffixed base form {base_form}
> Apply the following affixes to produce the final form:
> Affixes : {prefix} {suffix}
> Return ONE Arabic word only (no spaces, no punctuation).

**One-Shot Prompts:** We also test all the above prompts with the addition of a one-shot example. For instance, we add the following to the Root-Pattern prompt:

> Example (one-shot):
> Root: زرع | Template: {template} → Target form: {base_form(زرع)}
> Now answer for the requested root and pattern.

### 3.3.3. Evaluation

We evaluate model performance using generation accuracy. In practice, some language models tend to produce multiple-word outputs or additional context despite explicit prompt constraints. To account for this behavior, we adopt a lenient matching criterion: an output is considered correct if it contains a correctly formed word that conforms to the specified derivational pattern, even when embedded within a longer response.

## 4. Results & Analysis

Our results on token-morpheme alignment, along with probing task performance, are summarized in Tables 2 and 3. Overall, we find no consistent relationship between morpheme alignment scores and morphological generation accuracy. GPT4 and GPT4O exhibit opposite alignment patterns yet achieve the highest scores across all probing tasks. In contrast, the Arabic-centric models ALLAM and FANAR fail to generalize to nonce words, indicating reliance on lexeme memorization rather than true morphological generalization. In the following sections, we discuss these aspects in detail and conclude with broader insights and limitations.

### 4.1. Token-Morpheme Alignment Results

As can be seen from the bottom part of Table 2, morphological analyzers, namely CAMEL and FARASA, achieve fertility scores between 1 and 2, providing compression while preserving morphemes with high precision, recall, and MCR. This suggests that accurate morphemic segmentation can, in principle, be obtained without excessive splitting, serving as an oracle boundary, even if it is not practical for LLM development.

GPT4 exhibits clear over-segmentation (fertility > 3), producing many short tokens that artificially inflate boundary recall (85%), but this also results in very low precision (23%). The remaining tokenizers maintain fertility scores of approximately 1-2 similar to that in the morphological analyzers, indicating greater compression in input segmentation. At the morpheme level, most tokenizers (except GPT4) perform comparably, with ALLAM achieving the highest F1 scores. Their performance, however, remains well below that of morphological analyzers, as tokenizers are optimized for statistical efficiency rather than morpheme preservation. Boundary metrics reveal greater variation in how tokenizers align with morphemes: ALLAM, an Arabic-centric model lacking morpheme-based pre-tokenization, exhibit very low boundary recall. In contrast, FANAR also Arabic-centric but explicitly designed to preserve morphemes, achieves higher boundary recall, though its low precision (17%) and MCR suggest that it tends to over-segment morphemes. Multilingual tokenizers show similar precision to Arabic-centric ones, except for QWEN3 and GPT-4, which perform slightly better.

As shown in Figure 2, most tokenizers achieve higher morpheme F1 but lower boundary F1, while GPT4 exhibits the reverse trend. The figure also shows that morphological analyzers have inconsistent performance between MSA and dialectal data, being biased towards MSA, while all tokenizers show a consistent performance or a slightly better performance in dialectal data. This is likely a result of the distribution of data used to train these tokenizers, compared to the MSA-centric design of traditional morphological analyzers.

| Data | Model | Fertility | # Tokens | F1 | Boundary P | Boundary R | Boundary F1 | MCR |
|---|---|---|---|---|---|---|---|---|
| \multicolumn{9}{c}{**Tokenizers**} ||||||||
| ATB3 | ALLAM | 1.24 | 364,189 | **39.39** | 18.00 | 5.51 | 8.43 | **83.66** |
|  | FANAR | 1.91 | 561,373 | 34.32 | 17.96 | 20.61 | 19.19 | 49.46 |
|  | GPT4 | 4.01 | 1,175,386 | 24.59 | 23.07 | **85.21** | **36.31** | 33.04 |
|  | GPT4O | 1.76 | 516,062 | 35.28 | 17.64 | 16.83 | 17.22 | 56.44 |
|  | LLAMA3 | 2.15 | 631,730 | 32.85 | 18.75 | 27.15 | 22.18 | 42.62 |
|  | QWEN3 | 2.11 | 617,279 | 36.02 | **23.96** | 33.18 | 27.83 | 53.48 |
|  | COHERE | 1.92 | 564,098 | 33.30 | 19.33 | 22.38 | 20.74 | 49.65 |
| BOLT | ALLAM | 1.23 | 158,348 | **57.23** | **38.68** | 15.85 | 22.49 | **86.43** |
|  | FANAR | 1.77 | 227,951 | 43.99 | 26.23 | 35.63 | 30.22 | 55.26 |
|  | GPT4 | 3.33 | 428,143 | 21.28 | 22.90 | **92.40** | **36.70** | 26.66 |
|  | GPT4O | 1.69 | 216,801 | 45.17 | 26.54 | 32.01 | 29.02 | 60.38 |
|  | LLAMA3 | 1.94 | 248,937 | 39.01 | 23.00 | 37.81 | 28.60 | 46.38 |
|  | QWEN3 | 1.96 | 252,335 | 39.66 | 26.13 | 44.16 | 32.83 | 52.09 |
|  | COHERE | 1.84 | 237,063 | 40.15 | 23.77 | 35.22 | 28.38 | 50.35 |
| \multicolumn{9}{c}{**Morphological Analyzers**} ||||||||
| ATB3 | CAMEL | 1.49 | 437,563 | 73.24 | 98.41 | 60.94 | 75.27 | 97.07 |
|  | FARASA | 1.74 | 510,214 | **93.78** | **98.92** | **91.94** | **95.30** | **99.26** |
| BOLT | CAMEL | 1.27 | 163,073 | 66.17 | 76.60 | 36.32 | 49.27 | 85.29 |
|  | FARASA | 1.40 | 180,279 | **77.71** | **84.02** | **59.53** | **69.69** | **91.80** |

Table 2: Results of evaluating tokenizers and morphological analyzers. Evaluation metrics are described in §3.2.1. **Bold** values indicate the highest score for each dataset across tokenizers or analyzers.

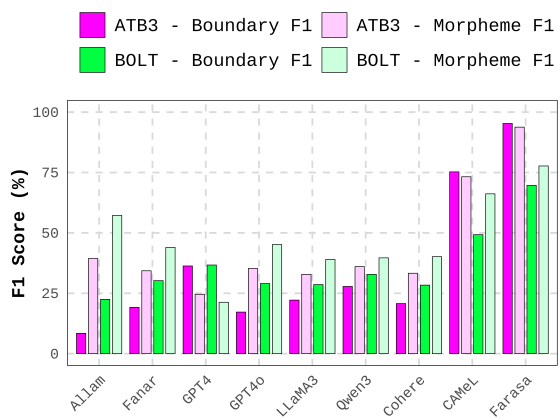

Figure 2: Comparison of Boundary F1 and Morpheme F1 scores on ATB3 and BOLT datasets.

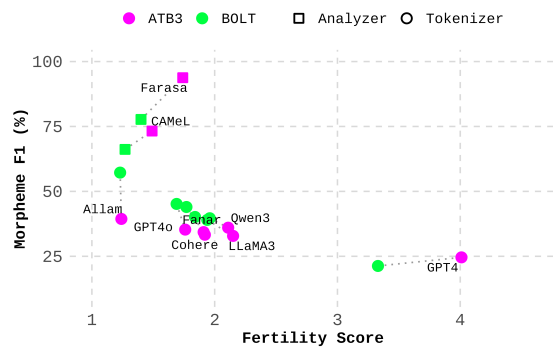

Figure 3: Fertility scores vs. morpheme F1 scores for tokenizers and analyzers on ATB3 and BOLT datasets.

Models show comparable fertility values but differ in their handling of morphemes, indicating that similar fertility does not ensure linguistic coherence. As shown in Figure 3, most tokenizers cluster within a narrow fertility range yet vary substantially in morphological fidelity.

**Estimation of Root Preservation.** Overall, AL-LAM exhibit strong morphological integrity compared to all other tokenizers as indicated by the high MCR score; it rarely splits morphemes internally, thus maintaining complete morphemes and roots. This behavior accounts for its low boundary recall (e.g., producing "wasa" instead of "wa sa"). All other tokenizers exhibit rather low MCR rates, with GPT4 being the lowest as a result of over-segmentation.

## 4.2. Morphological Productivity Results

As can be seen from Table 3, GPT4O consistently achieves the highest scores across all probing tasks. GPT4 performs similarly, demonstrating strong generalization despite its higher fertility. High performance on both real words (*Root-Pattern Real*) and nonce words (*Root-Pattern Nonce*) suggests that these models apply morphological transformation rules rather than rely on lexical mem-orization. In contrast, other models show clear limitations in productive morphology: their accuracies rarely exceed 60% and drop sharply on nonce words, reflecting possible over-reliance on memorized lexical patterns. The Arabic-centric models perform better than the multi-lingual models (other than the GPT4 series), which could be attributed to the size of Arabic training data. FANAR performs consistently across all probing tasks, while ALLAM shows a significant drop on nonce words. It is worth noting that FANAR incorporates morpheme-informed design, which may be a factor contributing to its steadier performance, but it could also be attributed to better instruction-following performance.

| Model | Root-Pattern (%) | | Affix Build (%) |
|---|---|---|---|
| | Real | Nonce | |
| ALLAM | 66.92 | 20.00 | 69.23 |
| FANAR | 56.92 | 52.00 | 44.23 |
| GPT4 | <u>94.62</u> | <u>92.00</u> | 88.46 |
| GPT4O | **96.92** | **97.00** | **91.92** |
| LLAMA3 | 26.15 | 10.00 | 68.08 |
| QWEN3 | 43.08 | 30.00 | 17.69 |
| COHERE | 43.85 | 29.00 | 60.00 |

Table 3: Generation accuracy across models. **Bold** indicates the highest score in each column, while <u>underlined</u> indicates the second highest. Results are shown for the English version of the 1-shot prompt, which yielded the highest or near-highest scores across most language models.

**A Note on Prompt Variations.** Table 3 presents the results for the English version of the 1-shot prompt. Before obtaining these scores, we explored several prompt variations to more effectively elicit the morphological transformation capabilities of the evaluated language models. The impact of one-shot prompting is evident in several models, which show clear improvements compared to their zero-shot counterparts.[5] These gains suggest that in-context examples help the models infer morphological relations more effectively. In contrast, GPT4 and GPT4O exhibit minimal differences across shot settings, indicating that they can apply the specified morphological rules without explicit exemplars. A cross-linguistic comparison reveals another pattern: most models perform substantially better with English prompts than Arabic, likely reflecting both the greater morphological complexity of Arabic and the relative scarcity of Arabic data in instruction fine-tuning.

---

[5]Comparison of zero-shot and one-shot prompts, and English vs. Arabic prompts, can be found in Table 5 in the Appendix.

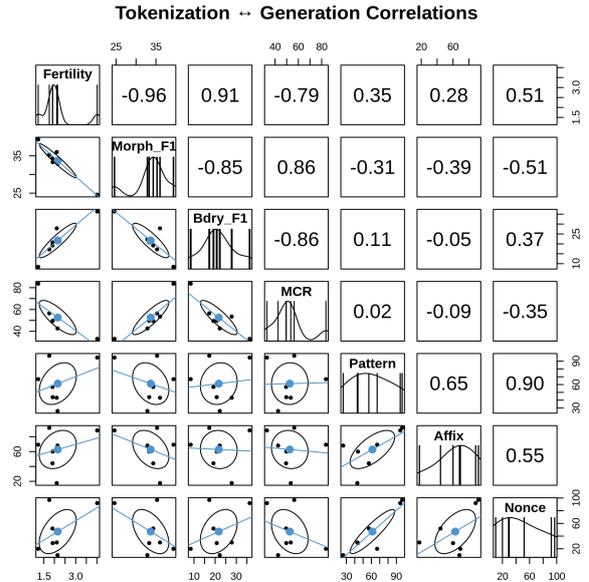

Figure 4: Pearson correlation between morphological alignment and generation scores.

### 4.3. Qualitative Analysis

Table 6 in the Appendix shows a breakdown of morphological generation accuracy across pattern types. We conducted a manual qualitative error analysis and examined outputs from each model, focusing on three patterns: فاعل /faʕil/, استفعل /istafʕala/, and فعال /fiʕaːl/. Table 4 summarizes the main error categories observed across models. Most models (except GPT4 and GPT4O) occasionally produce correct forms, but their overall behavior reflects partial, surface-level sensitivity to the instruction rather than consistent application of morphological rules.

### 4.4. Correlation with Internal Metrics.

As shown in Figure 4, we do not observe any consistent correlation between tokenizer alignment metrics such as boundary F1, or morpheme coverage rate, and model performance on the morphological probing or productivity tasks. Fertility scores are the only ones with consistent, but weak, positive correlation with generation tasks. This is likely due to the uniquely high fertility score of GPT4, which achieves the second highest generation accuracy in probing tasks. Strong morphological alignment metrics, such as Morpheme F1 or MCR, have zero or weak negative correlation with generation performance. These results suggest that surface tokenization quality in terms of morphological alignment is neither necessary nor sufficient for morphological productivity.

| | | Root–Pattern Errors | | | | Morpheme-Based Errors | |
|---|---|---|---|---|---|---|---|
| **Pattern Misapplication:** The correct root was identified but the pattern was misapplied. | GT<br>PRED<br>PATTERN<br>LLM | خفاص<br>خفص<br>فعال<br>Qwen | (/xifːaːs/)<br>(/xafas/)<br>(/fiʕaːl/) | **Missing Affix:** Model omits required prefix or suffix. | GT<br>PRED<br>PATTERN<br>LLM | جبالهم<br>جبال<br>فعال<br>Cohere | (/dʒibaːlhum/)<br>(/dʒibaːl/)<br>(/fiʕaːl/) |
| **Root Deformation:** Root consonants were altered (added, dropped, replaced). | GT<br>PRED<br>PATTERN<br>LLM | طغاد<br>غطاس<br>فعال<br>Allam | (/tˤɣaːd/)<br>(/ɣtˤaːs/)<br>(/fiʕaːl/) | **Wrong Base Selection:** Affixation is correct but added to an incorrect base word. | GT<br>PRED<br>PATTERN<br>LLM | العمال<br>العامل<br>فعال<br>Allam | (/alʕumaːl/)<br>(/alʕaːmil/)<br>(/fiʕaːl/) |
| **Wrong Pattern Selection:** A different morphological pattern was used instead of the target one. | GT<br>PRED<br>PATTERN<br>LLM | دغاز<br>داغز<br>فعال<br>Cohere | (/dɣaːz/)<br>(/daːɣz/)<br>(/fiʕaːl/) | **Wrong Morpheme Attachment:** Adds incorrect affixes instead of the intended ones. | GT<br>PRED<br>PATTERN<br>LLM | الطالبات<br>الطلاب<br>فاعل<br>Allam | (/altˤaːlibaːt/)<br>(/altˤulaːb/)<br>(/faːʕil/) |
| **Real Word Substitution:** Output is a valid Arabic word instead of applying the pattern to nonce roots. | GT<br>PRED<br>PATTERN<br>LLM | نفاغ<br>نافع<br>فعال<br>Allam | (/nifaːɣ/)<br>(/naːfiʕ/)<br>(/fiʕaːl/) | **Partial Truncation:** Emits only part of the target word. | GT<br>PRED<br>PATTERN<br>LLM | استمسك<br>استمس<br>استفعل<br>GPT4 | (/istamsak/)<br>(/istams/)<br>(/istafʕala/) |
| **Incorrect Affixation:** Affixes merged in incorrect order (prefix, root, infix misaligned). | GT<br>PRED<br>PATTERN<br>LLM | استخدم<br>خدماست<br>استفعل<br>LLaMA3 | (/istaxdam/)<br>(/xdmaːst/)<br>(/istafʕala/) | **Agreement / Category Error:** Incorrect number, gender, or case. | GT<br>PRED<br>PATTERN<br>LLM | الحالمين<br>الحالمون<br>فاعل<br>Allam | (/alħaːlimiːn/)<br>(/alħaːlimuːn/)<br>(/faːʕil/) |

Table 4: Common qualitative error types observed in Arabic morphological generation across evaluated models. GT = ground truth; PRED = model prediction; PATTERN = target morphological pattern; LM = language model.

## 5. Discussion & Implications

From our findings, we advocate a functional rather than structural view of morphological representation in LLMs. Morphological competence should be defined by a model's ability to generalize morphological transformations beyond the training vocabulary rather than focusing on surface morpheme segmentation. From this perspective, LLMs can exhibit morphological productivity even without aligned morpheme boundaries. There appears to be a level of subword segmentation that enables the model to represent words efficiently while remaining flexible enough to capture novel compositional structure within the framework of instruction tuning. Our results demonstrate that explicit morphological alignment in the tokenizer is neither necessary nor sufficient for productive generalization. Instead, compositional reasoning and instruction-following capabilities serve as functional substitutes for explicit morphological parsing, enabling rule-like transformations without predefined segmentation. Crucially, morpheme alignment does not predict morphological productivity: models with high alignment scores (e.g., Allam) do not achieve superior derivational performance, whereas GPT-4, despite its high fertility, generalizes more effectively. This challenges the assumption that morphology-aware tokenization is essential for modeling morphologically rich languages like Arabic.

The success of models like GPT4 in producing correct forms for nonce words in spite of having poor morphological alignment supports this hypothesis: productive patterns arise from statistical regularities in subword sequences rather than explicit and well-formed morphemes. Character-level interpolation and context-based reasoning allow models to manipulate internal consonantal templates even when token boundaries are poorly aligned with morphological units. Using instruction-following systems, morphology can be dynamically reconstructed rather than statically encoded. Practically, our results suggest that the role of linguistically informed tokenizers in large-scale LLM training may be questionable. Given the cost of developing language-specific tokenizers, future work could instead pursue tokenization-agnostic morphology learning, where LLMs acquire productive morphological rules through large-scale exposure and task-specific tuning. Adaptive or hybrid tokenization, allowing fine-grained character-level processing only when morphologically necessary, offers a promising direction.

## 5.1. Limitations

Our experimental setup can not directly probe deep morphological representation and reasoning and could instead conflate them with instruction-following ability, which could be what primarily drives morphological generation in our experiments. While we ensured that all evaluated models can follow the given instructions at least some of the time, the examined models vary in their instruction-following consistency, with some models generating unpredictably long text streams. To mitigate this effect, we tested multiple prompt formulations during probing and adopted a lenient evaluation criterion which accepts any correct target word among the top predictions, even when accompanied by other words. Considering the difficulty of tracing the cause of performance variations across models that have different architectures, training data, and optimization mechanisms, we based our analysis instead on correlation scores. However, we concede that such analysis is still limited and does not provide a basis for establishing causality. Our results should be used as a basis for further analysis and evaluation rather than a definitive conclusion about the relationship between tokenization and morphological generation. Controlled experiments examining the impact of tokenizer design on model performance and morphological representation could provide stronger insights into the causal relationship between tokenizer design and morphological productivity.

## 6. Bibliographical References

Ummar Abbas et al. 2025. Fanar: An arabic-centric multimodal generative ai platform. *arXiv preprint arXiv:2501.13944*.

Josh Achiam et al. 2024. Gpt-4 technical report. *arXiv preprint arXiv:2303.08774*.

Arash Ahmadian et al. 2025. Command a: An enterprise-ready large language model. *arXiv preprint arXiv:2504.00698*.

Mehdi Ali, Michael Fromm, Klaudia Thellmann, Richard Rutmann, Max Lübbering, Johannes Leveling, Katrin Klug, Jan Ebert, Niclas Doll, Jasper Buschhoff, et al. 2024. Tokenizer choice for llm training: Negligible or crucial? In *Findings of the Association for Computational Linguistics: NAACL 2024*, pages 3907–3924.

Catherine Arnett and Benjamin Bergen. 2025. Why do language models perform worse for morphologically complex languages? In *Proceedings of the 31st International Conference on Computational Linguistics*, pages 6607–6623, Abu Dhabi, UAE. Association for Computational Linguistics.

Catherine Arnett, Marisa Hudspeth, and Brendan O'Connor. 2025. Evaluating morphological alignment of tokenizers in 70 languages. In *Proceedings of the ICML 2025 Tokenization Workshop (TokShop)*. International Conference on Machine Learning.

Ehsaneddin Asgari, Yassine El Kheir, and Mohammad Ali Sadraei Javaheri. 2025. Morphbpe: A morpho-aware tokenizer bridging linguistic complexity for efficient llm training across morphologies. *arXiv preprint arXiv:2502.00894*.

M Saiful Bari, Yazeed Alnumay, Norah Alzahrani, Nouf Alotaibi, Hisham Alyahya, AlRashed Al-Rashed, Faisal Mirza, Shaykhah Alsubaie, Hassan Alahmed, Ghadah Alabduljabbar, Raghad Alkhathran, Yousef Almushayqih, Raneem Alnajim, Salman I Alsubaihi, Maryam Al Mansour, Saad Hassan, Majed Alrubaian, Ali Alammari, Zaki Alawami, Abdulmohsen Al-Thubaity, Ahmed Abdelali, Jeril Kuriakose, Abdalghani Abujabal, Nora Al-Twairesh, Areeb Alowisheq, and Haidar Khan. 2025. Allam: Large language models for arabic and english. In *International Conference on Learning Representations*, volume 2025, pages 34179–34214.

Kenneth R Beesley and Lauri Karttunen. 2003. Finite-state morphology: Xerox tools and techniques. *CSLI, Stanford*, pages 359–375.

Kaj Bostrom and Greg Durrett. 2020. Byte pair encoding is suboptimal for language model pretraining. In *Findings of the Association for Computational Linguistics: EMNLP 2020*, pages 4617–4624, Online. Association for Computational Linguistics.

Anh Dang, Limor Raviv, and Lukas Galke. 2024. Morphology matters: Probing the cross-linguistic morphological generalization abilities of large language models through a wug test. In *13th edition of the Workshop on Cognitive Modeling and Computational Linguistics (CMCL 2024)*, pages 177–188. Association for Computational Linguistics (ACL).

Kareem Darwish and Hamdy Mubarak. 2016. Farasa: A new fast and accurate arabic word segmenter. In *Proceedings of the Tenth International Conference on Language Resources and Evaluation (LREC'16)*, pages 1070–1074.

Philip Gage. 1994. A new algorithm for data compression. *C Users J.*, 12(2):23–38.

Bar Gazit, Shaltiel Shmidman, Avi Shmidman, and Yuval Pinter. 2025. Splintering nonconcatenative languages for better tokenization. In *Findings of the Association for Computational Linguistics: ACL 2025*, pages 22405–22417, Vienna, Austria. Association for Computational Linguistics.

Aaron Grattafiori et al. 2024. The llama 3 herd of models. *arXiv preprint arXiv:2407.21783*.

Nizar Y Habash. 2010. *Introduction to Arabic natural language processing*. Morgan & Claypool Publishers.

Valentin Hofmann, Hinrich Schuetze, and Janet Pierrehumbert. 2022. An embarrassingly simple method to mitigate undesirable properties of pretrained language model tokenizers. In *Proceedings of the 60th Annual Meeting of the Association for Computational Linguistics (Volume 2: Short Papers)*, pages 385–393, Dublin, Ireland. Association for Computational Linguistics.

Valentin Hofmann, Leonie Weissweiler, David R Mortensen, Hinrich Schütze, and Janet B Pierrehumbert. 2025. Derivational morphology reveals analogical generalization in large language models. *Proceedings of the National Academy of Sciences*, 122(19):e2423232122.

Binyuan Hui et al. 2025. Qwen2.5 technical report. *arXiv preprint arXiv:2409.12186*.

Aaron Hurst, Adam Lerer, Adam P Goucher, Adam Perelman, Aditya Ramesh, Aidan Clark, AJ Ostrow, Akila Welihinda, Alan Hayes, Alec Radford, et al. 2024. Gpt-4o system card. *arXiv preprint arXiv:2410.21276*.

Mete Ismayilzada, Defne Circi, Jonne Sälevä, Hale Sirin, Abdullatif Köksal, Bhuwan Dhingra, Antoine Bosselut, Duygu Ataman, and Lonneke Van Der Plas. 2025. Evaluating morphological compositional generalization in large language models. In *Proceedings of the 2025 Conference of the Nations of the Americas Chapter of the Association for Computational Linguistics: Human Language Technologies*. Association for Computational Linguistics.

Haris Jabbar. 2024. Morphpiece : A linguistic tokenizer for large language models. *arXiv preprint arXiv:2307.07262*.

Albert Q. Jiang, Alexandre Sablayrolles, Arthur Mensch, Chris Bamford, Devendra Singh Chaplot, Diego de las Casas, Florian Bressand, Gianna Lengyel, Guillaume Lample, Lucile Saulnier, Lélio Renard Lavaud, Marie-Anne Lachaux, Pierre Stock, Teven Le Scao, Thibaut Lavril, Thomas Wang, Timothée Lacroix, and William El Sayed. 2023. Mistral 7b.

Taku Kudo. 2018. Subword regularization: Improving neural network translation models with multiple subword candidates. In *Proceedings of the 56th Annual Meeting of the Association for Computational Linguistics (Volume 1: Long Papers)*, pages 66–75, Melbourne, Australia. Association for Computational Linguistics.

Sabrina J. Mielke, Zaid Alyafeai, Elizabeth Salesky, Colin Raffel, Manan Dey, Matthias Gallé, Arun Raja, Chenglei Si, Wilson Y. Lee, Benoît Sagot, and Samson Tan. 2021. Between words and characters: A brief history of open-vocabulary modeling and tokenization in nlp.

Ossama Obeid, Nasser Zalmout, Salam Khalifa, Dima Taji, Mai Oudah, Bashar Alhafni, Go Inoue, Fadhl Eryani, Alexander Erdmann, and Nizar Habash. 2020. CAMeL tools: An open source python toolkit for Arabic natural language processing. In *Proceedings of the Twelfth Language Resources and Evaluation Conference*, pages 7022–7032, Marseille, France. European Language Resources Association.

Artidoro Pagnoni, Ram Pasunuru, Pedro Rodriguez, John Nguyen, Benjamin Muller, Margaret Li, Chunting Zhou, Lili Yu, Jason Weston, Luke Zettlemoyer, et al. 2024. Byte latent transformer: Patches scale better than tokens. *arXiv preprint arXiv:2412.09871*.

Aleksandar Petrov, Emanuele La Malfa, Philip Torr, and Adel Bibi. 2023. Language model tokenizers introduce unfairness between languages. *Advances in neural information processing systems*, 36:36963–36990.

Phillip Rust, Jonas Pfeiffer, Ivan Vulić, Sebastian Ruder, and Iryna Gurevych. 2021. How good is your tokenizer? on the monolingual performance of multilingual language models. In *Proceedings of the 59th Annual Meeting of the Association for Computational Linguistics and the 11th International Joint Conference on Natural Language Processing (Volume 1: Long Papers)*, pages 3118–3135, Online. Association for Computational Linguistics.

Craig W Schmidt, Varshini Reddy, Chris Tanner, and Yuval Pinter. 2025. Boundless byte pair encoding: Breaking the pre-tokenization barrier. *arXiv preprint arXiv:2504.00178*.

Mike Schuster and Kaisuke Nakajima. 2012. Japanese and korean voice search. In *2012 IEEE International Conference on Acoustics, Speech and Signal Processing (ICASSP)*, pages 5149–5152.

Rico Sennrich, Barry Haddow, and Alexandra Birch. 2016. Neural machine translation of rare words

with subword units. In *Proceedings of the 54th Annual Meeting of the Association for Computational Linguistics (Volume 1: Long Papers)*, pages 1715–1725, Berlin, Germany. Association for Computational Linguistics.

Jean Seo, Jaeyoon Kim, SungJoo Byun, and Hyopil Shin. 2025. How does a language-specific tokenizer affect llms? *arXiv preprint arXiv:2502.12560*.

Leonie Weissweiler, Valentin Hofmann, Anjali Kantharuban, Anna Cai, Ritam Dutt, Amey Hengle, Anubha Kabra, Atharva Kulkarni, Abhishek Vijayakumar, Haofei Yu, Hinrich Schuetze, Kemal Oflazer, and David Mortensen. 2023. Counting the bugs in ChatGPT's wugs: A multilingual investigation into the morphological capabilities of a large language model. In *Proceedings of the 2023 Conference on Empirical Methods in Natural Language Processing*, pages 6508–6524, Singapore. Association for Computational Linguistics.

Leonie Weissweiler, Kyle Mahowald, and Adele Goldberg. 2025. Linguistic generalizations are not rules: Impacts on evaluation of lms. *arXiv preprint arXiv:2502.13195*.

Yonghui Wu, Mike Schuster, Zhifeng Chen, Quoc V. Le, Mohammad Norouzi, Wolfgang Macherey, Maxim Krikun, Yuan Cao, Qin Gao, Klaus Macherey, Jeff Klingner, Apurva Shah, Melvin Johnson, Xiaobing Liu, ukasz Kaiser, Stephan Gouws, Yoshikiyo Kato, Taku Kudo, Hideto Kazawa, Keith Stevens, George Kurian, Nishant Patil, Wei Wang, Cliff Young, Jason Smith, Jason Riesa, Alex Rudnick, Oriol Vinyals, Greg Corrado, Macduff Hughes, and Jeffrey Dean. 2016. Google's neural machine translation system: Bridging the gap between human and machine translation.

## 7. Language Resource References

Maamouri, Mohamed and Bies, Ann and Buckwalter, Tim and Mekki, Wigdan. 2010. *Arabic Treebank: Part 3 v 3.2 LDC2010T08*. ISLRN 770-467-034-042-0. Web Download. Philadelphia: Linguistic Data Consortium.

Maamouri, Mohamed and et al. 2021. *BOLT Egyptian Arabic Treebank - Conversational Telephone Speech LDC2021T12*. ISLRN 430-645-589-448-0. Web Download. Philadelphia: Linguistic Data Consortium.

# A. Additional Results

| Model | Prompt | Shots | Root-Pattern (%) | | Affix Build (%) |
|---|---|---|---|---|---|
| | | | Real | Nonce | |
| ALLAM | EN | 0 | 53.08 | **30.00** | 43.08 |
| ALLAM | EN | 1 | 66.92 | 20.00 | **69.23** |
| ALLAM | AR | 0 | 58.46 | 23.00 | 40.77 |
| ALLAM | AR | 1 | **80.77** | 15.00 | 58.85 |
| FANAR | EN | 0 | 31.54 | 14.00 | 37.69 |
| FANAR | EN | 1 | **56.92** | **52.00** | **44.23** |
| FANAR | AR | 0 | 0.77 | 2.00 | 36.15 |
| FANAR | AR | 1 | 32.31 | 21.00 | 7.31 |
| GPT4 | EN | 0 | 82.31 | 89.00 | 71.15 |
| GPT4 | EN | 1 | **94.62** | **92.00** | **88.46** |
| GPT4 | AR | 0 | 86.15 | 72.00 | 73.85 |
| GPT4 | AR | 1 | 84.62 | 78.00 | 31.54 |
| GPT4O | EN | 0 | 96.92 | 96.00 | 80.00 |
| GPT4O | EN | 1 | 96.92 | **97.00** | **91.92** |
| GPT4O | AR | 0 | 96.15 | 96.00 | 72.31 |
| GPT4O | AR | 1 | **97.69** | 95.00 | 90.77 |
| LLAMA3 | EN | 0 | **26.92** | 2.00 | 46.92 |
| LLAMA3 | EN | 1 | 26.15 | **10.00** | **68.08** |
| LLAMA3 | AR | 0 | 20.00 | 0.00 | 20.00 |
| LLAMA3 | AR | 1 | 16.15 | 7.00 | 57.69 |
| QWEN3 | EN | 0 | 26.15 | 16.00 | 12.31 |
| QWEN3 | EN | 1 | **43.08** | 30.00 | 17.69 |
| QWEN3 | AR | 0 | 21.54 | 5.00 | 23.46 |
| QWEN3 | AR | 1 | 37.69 | **40.00** | **67.69** |
| COHERE | EN | 0 | 37.69 | 16.00 | 35.77 |
| COHERE | EN | 1 | **43.85** | **29.00** | **60.00** |
| COHERE | AR | 0 | 21.54 | 3.00 | 11.54 |
| COHERE | AR | 1 | 7.69 | 11.00 | 28.46 |

Table 5: Generation accuracy across models and prompting settings (zero-shot vs. one-shot / English vs. Arabic prompts). **Bold** indicates the highest score in each column per model.

| Pattern | Model | Root-Pattern Real | Root-Pattern Nonce | Affix |
|---|---|---|---|---|
| مفعول | Allam | 70 | 10 | 70 |
| | Fanar | 80 | 40 | 40 |
| | GPT-4 | **100** | 80 | **90** |
| | GPT-4o | **100** | **95** | **90** |
| | LLaMA-3 | 10 | 5 | 60 |
| | Qwen-3 | 60 | 5 | 40 |
| | Cohere | 30 | 15 | 50 |
| فاعل | Allam | **100** | 30 | 80 |
| | Fanar | 90 | 45 | 10 |
| | GPT-4 | **100** | **100** | **100** |
| | GPT-4o | **100** | **100** | **100** |
| | LLaMA-3 | 50 | 20 | 75 |
| | Qwen-3 | 30 | 20 | 5 |
| | Cohere | 70 | 25 | 50 |
| فعالة | Allam | 70 | ... | 45 |
| | Fanar | 50 | ... | 30 |
| | GPT-4 | **90** | ... | 80 |
| | GPT-4o | **90** | ... | **85** |
| | LLaMA-3 | 60 | ... | 55 |
| | Qwen-3 | 50 | ... | 0 |
| | Cohere | 40 | ... | 60 |
| استفعل | Allam | 20 | 40 | **100** |
| | Fanar | **100** | 75 | 55 |
| | GPT-4 | **100** | **95** | **100** |
| | GPT-4o | **100** | **95** | **100** |
| | LLaMA-3 | 0 | 10 | 90 |
| | Qwen-3 | 80 | 85 | 10 |
| | Cohere | 90 | 55 | 30 |
| فعيل | Allam | 80 | ... | 60 |
| | Fanar | 40 | ... | 75 |
| | GPT-4 | **100** | ... | **95** |
| | GPT-4o | **100** | ... | 90 |
| | LLaMA-3 | 40 | ... | 55 |
| | Qwen-3 | 50 | ... | 15 |
| | Cohere | 50 | ... | 70 |
| فعلان | Allam | 90 | ... | 45 |
| | Fanar | 80 | ... | 30 |
| | GPT-4 | **100** | ... | **70** |
| | GPT-4o | **100** | ... | **70** |
| | LLaMA-3 | 40 | ... | 50 |
| | Qwen-3 | 90 | ... | 10 |
| | Cohere | 50 | ... | 45 |
| مفعال | Allam | 30 | ... | 60 |
| | Fanar | 50 | ... | 35 |
| | GPT-4 | **90** | ... | 95 |
| | GPT-4o | **90** | ... | **100** |
| | LLaMA-3 | 30 | ... | 90 |
| | Qwen-3 | 20 | ... | 20 |
| | Cohere | 30 | ... | 85 |

(a) Patterns 1–7

| Pattern | Model | Root-Pattern Real | Root-Pattern Nonce | Affix |
|---|---|---|---|---|
| انفعل | Allam | 80 | ... | 90 |
| | Fanar | 60 | ... | 75 |
| | GPT-4 | **100** | ... | 70 |
| | GPT-4o | **100** | ... | **100** |
| | LLaMA-3 | 0 | ... | 75 |
| | Qwen-3 | 50 | ... | 0 |
| | Cohere | 60 | ... | 50 |
| مفتعل | Allam | 60 | ... | 75 |
| | Fanar | 20 | ... | 55 |
| | GPT-4 | 50 | ... | 90 |
| | GPT-4o | **100** | ... | **95** |
| | LLaMA-3 | 10 | ... | 90 |
| | Qwen-3 | 0 | ... | 25 |
| | Cohere | 0 | ... | 50 |
| افتعال | Allam | 40 | ... | 85 |
| | Fanar | 20 | ... | 80 |
| | GPT-4 | **100** | ... | 90 |
| | GPT-4o | 80 | ... | **95** |
| | LLaMA-3 | 0 | ... | 60 |
| | Qwen-3 | 10 | ... | 30 |
| | Cohere | 10 | ... | 65 |
| فعول | Allam | **100** | 15 | 75 |
| | Fanar | 20 | 35 | 30 |
| | GPT-4 | **100** | 90 | **100** |
| | GPT-4o | **100** | **100** | **100** |
| | LLaMA-3 | 30 | 5 | 50 |
| | Qwen-3 | 10 | 25 | 25 |
| | Cohere | 50 | 35 | 85 |
| فعال | Allam | 60 | 5 | 70 |
| | Fanar | 50 | 65 | 45 |
| | GPT-4 | **100** | **95** | **100** |
| | GPT-4o | **100** | **95** | **100** |
| | LLaMA-3 | 30 | 10 | 75 |
| | Qwen-3 | 20 | 15 | 25 |
| | Cohere | 60 | 15 | 85 |
| فعلاء | Allam | 70 | ... | 45 |
| | Fanar | 80 | ... | 15 |
| | GPT-4 | **100** | ... | **70** |
| | GPT-4o | **100** | ... | **70** |
| | LLaMA-3 | 40 | ... | 60 |
| | Qwen-3 | 90 | ... | 25 |
| | Cohere | 30 | ... | 55 |

(b) Patterns 8–13

Table 6: Generation accuracy obtained with English one-shot prompts across 13 Arabic derivational patterns across models. "..." indicates that this pattern is not present for the nonce words in our dataset. **Bold** indicates the highest score for each column per model.